\documentclass[10pt,twocolumn,twoside]{IEEEtran}
\IEEEoverridecommandlockouts

\usepackage{cite}
\usepackage{amsmath,amssymb,amsfonts}
\usepackage{algorithmic}
\usepackage{graphicx}
\usepackage{textcomp}
\usepackage{balance}
\usepackage{xcolor}
\usepackage[utf8]{inputenc}
\usepackage{hyperref}
\usepackage{booktabs}
\usepackage{multirow}
\usepackage{tablefootnote}
\usepackage{wrapfig}

\def\BibTeX{{\rm B\kern-.05em{\sc i\kern-.025em b}\kern-.08em
    T\kern-.1667em\lower.7ex\hbox{E}\kern-.125emX}}
\usepackage{lipsum}

\graphicspath{ {./images/} }

\title{\LARGE \bf 
Generating adversarial inputs for a graph neural network model of AC power flow
}

\author{Robert Parker$^\dagger$
\thanks{$^\dagger$Los Alamos National Laboratory, Los Alamos, NM, USA\\ LA-UR-26-20748}
}

\newcommand{\PF}{\mathrm{PF}}
\newcommand{\NN}{\mathrm{NN}}

\begin{document}
\begingroup
\allowdisplaybreaks

\maketitle

\begin{abstract}
This work formulates and solves optimization problems to generate input points
that yield high errors between a neural network's predicted AC power flow
solution and solutions to the AC power flow equations. We demonstrate this
capability on an instance of the CANOS-PF graph neural network model,
as implemented by the PF$\Delta$ benchmark library, operating on a 14-bus test grid.
Generated adversarial points yield errors as large as 3.7 per-unit
in reactive power and 0.08 per-unit in voltage magnitude.
When minimizing the perturbation from a training point necessary to
satisfy adversarial constraints, we find that the constraints can be
met with as little as an 0.04 per-unit perturbation in voltage magnitude
on a single bus.
This work motivates the development of rigorous verification and robust
training methods for neural network surrogate models of AC power flow.
\end{abstract}


\vspace{-0.2cm}
\section{Introduction}
Recent work has focused on developing neural network surrogate models to approximate
solutions to the alternating current power flow (ACPF) equations.
Different architectures such as physics-informed neural networks (PINNs)
\cite{nellikkath2022,jalving2024physics}
and graph neural networks (GNNs) \cite{donon2020gns,lin2024pfnet,piloto2024canos}
have been proposed
and datasets such as OPF-Learn \cite{opflearn}, OPFData \cite{OPFData}, and
PF$\Delta$ \cite{pfdelta} have been developed to standardize comparisons among
different neural (optimal) power flow solvers.
These neural network models achieve fast online inference in exchange for
a (usually) small approximation error and large offline training cost.
Fast online inference is advantageous in the power transmission setting where
dispatch decisions are made frequently and network topologies are generally static.
However, it is well-known that neural networks are not robust to adversarial perturbations
\cite{szegedy2014intriguing,wiyatno2019adversarial}.
In safety-critical fields such as electric power transmission, robustness of
neural network-based algorithms is especially important.
This work addresses the following question: Given a particular surrogate model
for AC power flow, can we find input points that have large errors between
the surrogate's output and a ``ground truth'' solution of the AC power flow
equations?

Our contribution is to formulate and solve two types of optimization problems
for generating adversarial input points for an instance of the CANOS-PF
graph neural network ACPF surrogate operating on the IEEE 14-bus
test system from PGLib-OPF \cite{pglib} and trained on data from PF$\Delta$.
(See \cite{piloto2024canos} for details on the CANOS
  architecture; we use the open-source CANOS-PF GNN model implemented by
PF$\Delta$ \cite{pfdelta}.)
We demonstrate that these optimization problems are tractable for a
state-of-the-art embedded neural network model and that, for this model,
there exist adversarial points that introduce significant discrepancies
between the NN and ACPF solutions. These points
introduce large deviations in output variables that are
critical for grid operators to predict accurately, such as reactive
powers and voltage magnitudes. These results motivate further research
into verification and robust training methods for neural network surrogates
of AC power flow.

\vspace{-0.2cm}
\subsection{Related work}
Many works have addressed the issue of surrogate model robustness in power system operation.
Chen et al. \cite{chen2018vulnerable} have identified adversarial inputs that result
in misclassifications and forecasting errors, while Dinh et al. \cite{dinh2023reliability}
have analyzed the input points that cause a multi-layer perceptron (MLP) model to make
inaccurate predictions. Chevalier et al. have identified loading conditions that cause
(1) a DCOPF solution to be AC-infeasible \cite{chevalier2026adversarial} and (2) an MLP's
line switching decisions to cause high load shedding \cite{chevalier2025maximal}.
Simultaneously, methods for verification \cite{jalving2024physics},
robust training \cite{giraud2025training},
and guaranteeing feasibility at inference time \cite{donti2021dc}
of neural networks for AC power flow have been developed.
In contrast to the ``adversarial input'' papers above, our work targets a graph neural network
model that is state-of-the-art (according to a recent benchmark \cite{pfdelta}).
We also impose constraints on the neural network's outputs (see Problem \ref{eqn:con-error}),
which has only been done by \cite{chevalier2025maximal}.

\section{Problem formulation}
We formulate and solve two types of optimization problems:
\textit{Maximum-error} problems that maximize the difference between
specified coordinates of the neural network's output and ACPF solution
and \textit{constrained-error} problems that find the smallest input perturbations
that satisfy constraints on each output.
Each problem instance targets a specific output, such as voltage magnitude on a PV bus.
In constrained-error problems, we constrain a particular output
to be ``sufficiently different'' between the NN and ACPF solutions.

The maximum-error problem is given by Problem \ref{eqn:max-error}:
\begin{equation}
    \max~ \left( y_{\mathrm{NN},i} - y_{\mathrm{PF},i}\right)
    \text{ such that}~ \left\{\begin{array}{l}
        y_\mathrm{NN} = \mathrm{NN}(x) \\
        y_\PF = \PF(x) \\
        L\leq x \leq U \\
 \end{array}\right.
  \label{eqn:max-error}
\end{equation}
Here, $x$ is a vector of inputs to the AC power flow equations: Active power injections and voltage
magnitudes on PV buses, active and reactive powers on PQ buses, and voltage angle and magnitude
on the reference bus.
Function $\NN$ evaluates the neural network. Vector $y_\NN$ contains the outputs predicted
by the neural network: Reactive powers and voltage angles at PV buses, voltage angles and magnitudes
at PQ buses, and active and reactive power at the reference bus.
Function $\PF$ solves the AC power flow equations and returns $y_\PF$, the same outputs, now determined by
these equations. Bounds on the input vector ensure this problem remains bounded.
Problem \ref{eqn:max-error} maximizes the difference between the $i$-th coordinate of
the neural network's output and the $i$-th coordinate of the AC power flow equations' output.
We solve instances of Problem \ref{eqn:max-error} with both maximization and minimization objective
senses to maximize both the positive and negative errors between the neural network's output
and the power flow equations' output.

The constrained-error problem is given by Problem \ref{eqn:con-error}:
\begin{equation}
  \min~ \left\|x - x_0\right\|_1 \text{ such that} ~\left\{\begin{array}{ll}
      y_\NN = \NN(x); & y_{\NN,i} \geq l_i \\
      y_\PF = \PF(x); & y_{\PF,i} \leq l_i - \delta\\
    L\leq x \leq U \\
  \end{array}\right.
  \label{eqn:con-error}
\end{equation}
Vector $x_0$ is a reference point obtained from the training data of our
surrogate model. We use constraints to enforce a minimum difference
specified by the constant parameter $\delta$
between the $i$-th coordinate of the neural network and power flow output
vectors. Constant parameter $l_i$ emulates a lower bound that would typically
be applied to this output in an optimal power flow problem, such as a lower
bound on a voltage magnitude output. Adversarial points identified by
Problem \ref{eqn:con-error} can be interpreted as points where the
neural network predicts a feasible output (according to the bound $l_i$
in the selected coordinate)
but solving the AC power flow equations yields a solution that violates
this bound by a margin of at least $\delta$.
While we define this problem as targeting a single lower bound $l_i$, the problem
could easily be adjusted to target any number of lower or upper bounds.

\vspace{-0.2cm}
\section{Test problem}
We solve instances of Problems \ref{eqn:max-error} and \ref{eqn:con-error}
with an implementation of the CANOS-PF graph neural network operating on a
14-bus test grid. We use the hyperparameters suggested by \cite{pfdelta}:
A hidden dimension of 128, 15 message passing steps, a learning rate of
$10^{-5}$, and 50 training epochs. The model has approximately
7.5 million trainable parameters.

We use the IEEE 14-bus test network from PGLib-OPF \cite{pglib}. Input bounds are
taken from the default PGLib test case. In the constrained-error problem,
our output constraints target voltage magnitude at a specified PQ bus.
We use the lower bound provided by the PGLib test case and a margin
of $\delta=0.04$ (per-unit).
While the input vector, $x$, contains net loads on PQ buses, we use
the convention that
loads are non-dispatchable and are therefore fixed to their nominal values
(the default PGLib values in the case of Problem \ref{eqn:max-error} and
the values from the training data point $x_0$ in the case of Problem \ref{eqn:con-error}).
Similarly, we fix inputs corresponding to reference bus voltage angle and magnitude
to zero and 1.0 (per-unit).
This significantly constrains the feasible space of these optimization problems.
Degrees of freedom are net active power injections $p^{\rm net}$ and voltage magnitudes $v$ at
PV buses.

\section{Results}
We use PF$\Delta$'s CANOS-PF model implementation, dataset,
and training script. We use the Task-1.1 training dataset,
which provides 48,600 input and output points representing solved
power flow instances on the full network topology (no contingencies).
We implement the optimization problem 
using JuMP \cite{jump1} and MathProgIncidence.jl \cite{parker2023dulmage}.
PowerModels \cite{coffrin2018powermodels} implements the AC power flow
constraints and MathOptAI.jl \cite{mathoptai} implements the neural network constraints
by interfacing with PyTorch \cite{paszke2019pytorch} via a GPU-accelerated gray-box formulation
\cite{parker2025gpu}.
We solve optimization problems (locally) with IPOPT \cite{ipopt} using MA57 \cite{ma57} as the linear
solver, a tolerance of $10^{-6}$, an ``acceptable tolerance'' of $10^{-4}$, and
an iteration limit of 500. All computational experiments, including neural
network training, were performed on the Selene supercomputer at Los Alamos National Laboratory
using an Intel 8470 CPU and NVIDIA H100 GPU. The code used to produce these results is
available at \url{https://github.com/Robbybp/pfdelta}.

\vspace{-0.3cm}
\subsection{Neural network training results}
We train the CANOS-PF model with the PF$\Delta$ Task-1.1 training dataset and script
for 50 epochs with a learning rate of $5\times 10^{-4}$.
Training uses a combination of mean-squared-error (MSE) loss, a supervised
measure, and ``constraint violation loss,'' an unsupervised measure that
penalizes violation of the AC power flow equations. See \cite{fioretto2020}
for details behind this combined loss and \cite{piloto2024canos} for the
particular implementation in CANOS.
Following \cite{pfdelta}, we evaluate the quality of our trained neueral network
with MSE loss and ``power balance loss'' (PBL),
an unsupervised measure that penalizes only violation of active and reactive
power balance constraints.
\begin{table}
  \centering
  \caption{Losses obtained by our trained CANOS-PF model}
  \resizebox{\columnwidth}{!}{
  \begin{tabular}{ccccccccc}
    \toprule
    \multirow{2}{*}{Dataset} & \multirow{2}{*}{N. points} & \multicolumn{3}{c}{MSE} & \multicolumn{3}{c}{PBL} \\
    \cmidrule(lr){3-5}\cmidrule(lr){6-8}
    & & Mean & Stdev. & Max. & Mean & Stdev. & Max. \\
    \midrule
    Train & 48,600 & 3e-4 & 2e-5 & 4e-4 & 1.6e-2 & 2e-4 & 1.7e-2 \\
    Adversarial
    & 87 & 0.7 & 1.0 & 5.7 & 0.3 & 0.1 & 0.6 \\
    \bottomrule
  \end{tabular}
}
  \label{tab:loss}
\end{table}
%
%

Loss values on train and test data are shown in Table \ref{tab:loss}.
These values are comparable to MSE and PBL losses reported by
\cite{pfdelta}, so we conclude that we have accurately reproduced
this work's CANOS-PF model.
(Compare our training MSE loss to Table A.7 and our PBL to
Table A.8 in \cite{pfdelta}. While neither is a perfect comparison,
our trained model's loss is lower than those presented in \cite{pfdelta}
in both categories, as expected for \textit{training} loss and our smaller network.)
The ``Adversarial'' dataset in Table \ref{tab:loss} refers to the adversarial
points generated by solving Problems \ref{eqn:max-error} and \ref{eqn:con-error}:
23 points from Problem \ref{eqn:max-error} and 64 points from Problem \ref{eqn:con-error}.
The generation of these points is discussed in Sections \ref{sec:max-error}
and \ref{sec:con-error}.
\begin{figure}[h]
  \centering
  \includegraphics[width=4cm]{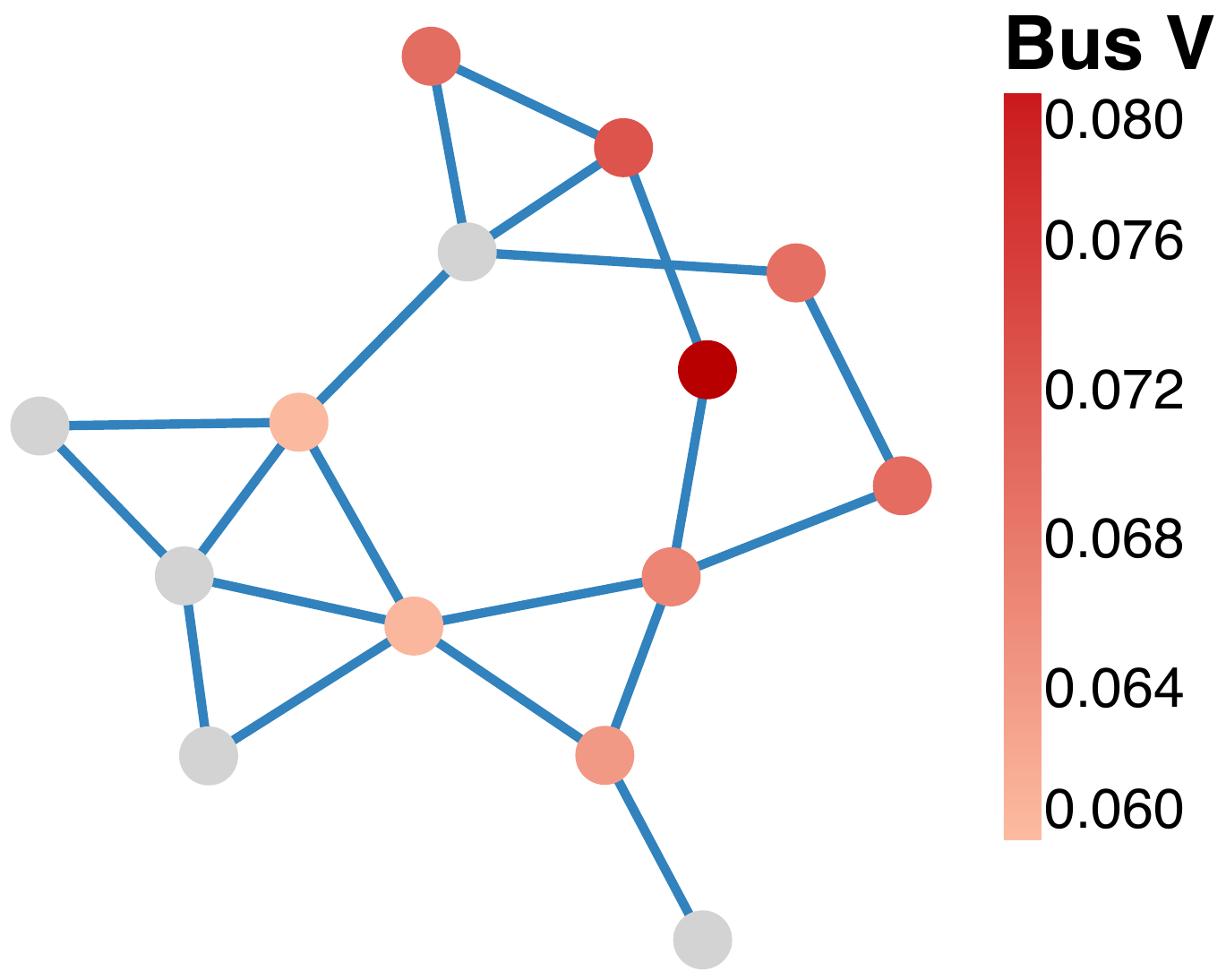}
  \includegraphics[width=4cm]{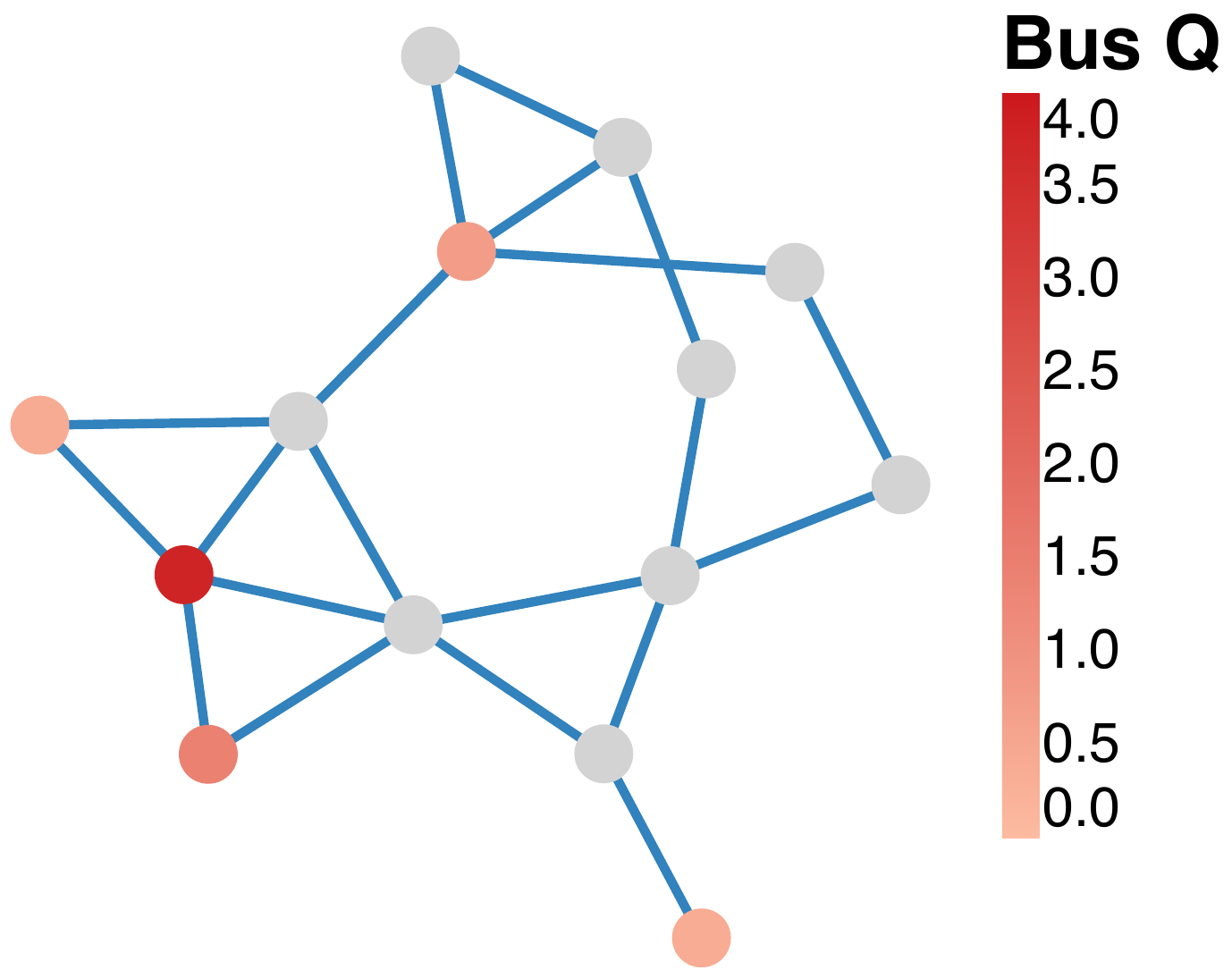}
  \caption{Maximum absolute error obtained for each bus. Left: PQ buses.
  Right: PV and reference buses. In each case, the opposite set of buses
  is shaded gray. Generated using PowerPlots \cite{powerplots}.}
  \label{fig:max-error-grid}
\end{figure}

\vspace{-0.2cm}
\subsection{Maximum-error results}
\label{sec:max-error}
We solve two instances of Problem \ref{eqn:max-error} for each bus: One
maximizing $(y_{\NN,i}-y_{\PF,i})$ and one minimizing the same quantity.
For PQ buses, the target output (coordinate $i$) is voltage magnitude, $v$;
for PV and reference buses, the target output is reactive power, $q$.
These errors we achieve are illustrated in Figure \ref{fig:max-error-grid},
where the maximum absolute values between maximization and minimization
errors are displayed.
The results for both maximization and minimization problems are displayed
in Figure \ref{fig:max-error-bar}.
\begin{figure}[h]
  \centering
  \vspace{-0.7cm}
  \includegraphics[width=8cm]{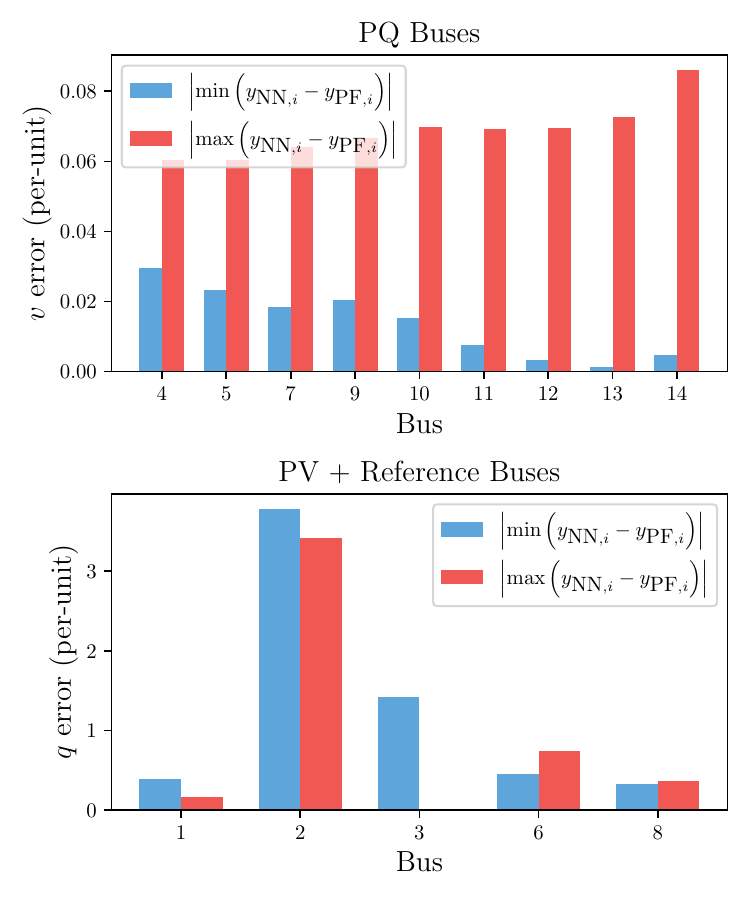}
  \vspace{-0.5cm}
  \caption{Errors achieved by Problem \ref{eqn:max-error} with maximization
  and minimization objectives. Bars are omitted where the solver
fails to converge.}
  \vspace{-0.4cm}
  \label{fig:max-error-bar}
\end{figure}

The results indicate that large errors between the neural network and AC
power flow solutions can be found. Voltage magnitude errors appear to be
systematically larger when maximizing than when minimizing, suggesting
that the neural network has difficulty approximating solutions accurately
when the power flow solution yields a relatively low voltage.


\vspace{-0.3cm}
\subsection{Constrained-error results}
\label{sec:con-error}
We solve instances of Problem \ref{eqn:con-error} for each of nine PQ buses and
for the first ten training points of the PF$\Delta$ Task-1.1 dataset.
Table \ref{tab:perturbation-summary} summarizes the results by training point.
We first note that all PQ buses have voltage magnitude lower bounds (according to
PGLib) of 0.94, so our constraint on $y_{\PF,i}$ (with margin $\delta=0.04$) is
always $y_{\PF,i}\leq 0.90$. The errors in voltage magnitude we achieve range
from 0.04 to 0.06.
There is a large variation in the number of successfully converged problems
for different training points, suggesting that some training points are
more vulnerable to adversarial perturbations than others.
Overall, 64 / 90 instances converge to primal-feasible points.

\begin{table}
  \centering
  \caption{Summary of adversarial perturbations by training point}
  \resizebox{\columnwidth}{!}{
  \begin{tabular}{cccccc}
    \toprule
    Training point & Converged & \multicolumn{4}{c}{Average} \\
    index & (of 9) & $\left\|x-x_0\right\|_1$ & $\left\|x-x_0\right\|_0$ & $y_{\NN,i}$ & $y_{\PF,i}$ \\
    \midrule
    0 & 9 & 0.31 & 3 & 0.96 & 0.90 \\
    1 & 4 & 0.29 & 3 & 0.96 & 0.90 \\
    2 & 8 & 0.14 & 3 & 0.96 & 0.90 \\
    3 & 6 & 0.11 & 2 & 0.96 & 0.90 \\
    4 & 6 & 0.10 & 2 & 0.96 & 0.90 \\
    5 & 9 & 0.21 & 2 & 0.95 & 0.90 \\
    6 & 8 & 0.13 & 2 & 0.95 & 0.90 \\
    7 & 6 & 0.26 & 3 & 0.95 & 0.90 \\
    8 & 3 & 0.09 & 2 & 0.94 & 0.90 \\
    9 & 5 & 0.18 & 2 & 0.96 & 0.90 \\
\bottomrule
  \end{tabular}
}
  \label{tab:perturbation-summary}
\end{table}

\begin{table}[h]
  \centering
  \caption{Adversarial perturbations for selected cases}
  \resizebox{\columnwidth}{!}{
  \begin{tabular}{ccccc}
    \toprule
    Training & Targeted & \multirow{2}{*}{$\left\|x-x_0\right\|_1$} & \multirow{2}{*}{$\left\|x - x_0\right\|_0$} & \multirow{2}{*}{Variable values} \\
  point & bus \\
  \midrule
  4 & 12 & 0.04 & 1 & $v_6=0.97$ \\[0.2cm]
  0 & 4  & 0.18 & 2 & $v_2=0.94,~v_3=0.95$ \\[0.2cm]
  2 & 5  & 0.10 & 3 & $\begin{array}{c} v_1=0.94, ~v_2=0.94, \\ v_6=1.03 \end{array}$\\[0.4cm]
  7 & 7  & 0.41 & 4 & $\begin{array}{c} v_2=0.94, ~v_3=0.95, \\ v_6=0.94, ~v_8=0.94 \end{array}$\\[0.4cm]
  0 & 13 & 1.03 & 6 & $\begin{array}{c} v_1=0.94, ~p_2^\mathrm{net}=-0.02 \\ v_2=0.94, ~v_3=0.94, \\ v_6=0.94, ~v_8=0.94 \end{array}$\\
  \bottomrule &
  \end{tabular}
}
  \label{tab:perturbations}
  \vspace{-0.5cm}
\end{table}

The ``0-norm'' in Table \ref{tab:perturbation-summary} is the number of nonzero
entries of its argument, i.e., the number of input variables
that need to be perturbed to achieve the adversarial outcome.
\begin{wrapfigure}{l}{0.55\columnwidth}
  \includegraphics[width=0.55\columnwidth]{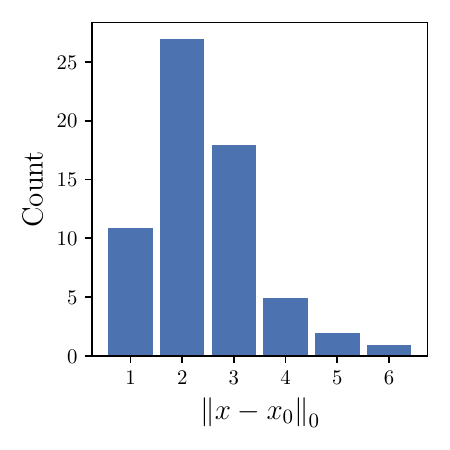}
  \vspace{-0.8cm}
  \caption{Histogram of 0-norm of perturbations necessary to satisfy adversarial constraints
  for the 64 converged instances of Problem \ref{eqn:con-error}.}
  \vspace{-0.2cm}
  \label{fig:l0-histogram}
\end{wrapfigure}
We note that these numbers are low. As shown by the histogram in Figure \ref{fig:l0-histogram},
the adversarial outcome is achieved by perturbing two or fewer input variables in most cases.
Selected adversarial perturbations are shown in Table \ref{tab:perturbations}.
We note that each variable listed corresponds to a single perturbed coordinate
of $x$.
As shown in Table \ref{tab:perturbations}, these perturbations
often require setting voltage magnitudes on PV buses to values close to their lower bounds of 0.94.
This result suggests that robustness of the CANOS-PF neural network model could be improved
by including more points with output voltages near this bound in the training data.

\vspace{-0.3cm}
\begin{table}
  \centering
  \caption{Statistics from optimization problem solves}
  \resizebox{\columnwidth}{!}{
  \begin{tabular}{cccccccc}
    \toprule
    Problem & Converged & Var. & Con. & JNNZ & HNNZ & Iter. & Time (s) \\
    \midrule
    (\ref{eqn:max-error}) & 23 / 28 & 470 & 481 & 30k & 24k & 168  & 92 \\
    (\ref{eqn:con-error}) & 64 / 90 & 902 & 699 & 30k & 24k & 25 & 14 \\
    \bottomrule
  \end{tabular}
}
  \label{tab:opt}
\end{table}

\subsection{Computational details}
Table \ref{tab:opt} reports information about the optimization problems solved.
Iteration counts and solve times are averages over the converged cases
where ``converged'' is defined as terminating at a primal-feasible point within 500 iterations.
Computational cost is
dominated by function and derivative evaluations, which we assume is due to
the cost of evaluating and differentiating the CANOS-PF model at every iteration.
These solve times are relatively high despite offloading CANOS-PF evaluation to
a GPU.
Fast optimization solves with large, dense, and sequential NNs embedded have been demonstrated
\cite{parker2025gpu}. However, it is not clear whether these results should extend to the
more intricate CANOS GNN architecture. Scalability of Problems \ref{eqn:max-error}
and \ref{eqn:con-error} will depend on how efficiently GNNs such as CANOS can be differentiated
at larger scales.

Finally, we note that the CANOS model violates the assumptions of the IPOPT solver due to its
many non-differentiable ReLU layers. It is surprising that we can converge 74\% of the problems
attempted despite this apparent incompatibility.

\section{Conclusion}
This work is a proof-of-concept that adversarial points can be systematically identified
for a state-of-the-art neural network model of AC power flow. We have demonstrated the method
on a small, 14-bus test system.
Future work should apply these methods to larger networks and additional NN models.
We note that by keeping loads fixed, we have significantly constrained the input
space that we search for adversarial points. Worse points than those shown here can likely
be generated if loads (and line parameters) are used as degrees of freedom as well.
This work motivates the need for continued robustness testing of neural network surrogate
models for AC power flow, development of rigorous validation methods for these models,
and development of adversarially robust training and inference methodologies.

\section*{Acknowledgements}
AI was used to write scripts to generate some of the figures and tables presented in
this work after the raw data had been generated by a human programmer.


\bibliographystyle{ieeetr}
\bibliography{ref}

@misc{piloto2024canos,
      title={{CANOS}: A Fast and Scalable Neural {AC-OPF} Solver Robust To {N-1} Perturbations}, 
      author={Luis Piloto and Sofia Liguori and Sephora Madjiheurem and Miha Zgubic and Sean Lovett and Hamish Tomlinson and Sophie Elster and Chris Apps and Sims Witherspoon},
      year={2024},
      eprint={2403.17660},
      archivePrefix={arXiv},
      primaryClass={cs.LG},
      url={https://arxiv.org/abs/2403.17660}, 
}

@article{lin2024pfnet,
title = {{PowerFlowNet}: {Power} flow approximation using message passing Graph Neural Networks},
journal = {International Journal of Electrical Power \& Energy Systems},
volume = {160},
pages = {110112},
year = {2024},
issn = {0142-0615},
doi = {https://doi.org/10.1016/j.ijepes.2024.110112},
author = {Nan Lin and Stavros Orfanoudakis and Nathan Ordonez Cardenas and Juan S. Giraldo and Pedro P. Vergara},
}

@article{jalving2024physics,
title = {Physics-informed machine learning with optimization-based guarantees: {Applications} to {AC} power flow},
journal = {International Journal of Electrical Power \& Energy Systems},
volume = {157},
pages = {109741},
year = {2024},
issn = {0142-0615},
doi = {https://doi.org/10.1016/j.ijepes.2023.109741},
author = {Jordan Jalving and Michael Eydenberg and Logan Blakely and Anya Castillo and Zachary Kilwein and J. Kyle Skolfield and Fani Boukouvala and Carl Laird},
}

@article{donon2020gns,
title = {Neural networks for power flow: Graph neural solver},
journal = {Electric Power Systems Research},
volume = {189},
pages = {106547},
year = {2020},
issn = {0378-7796},
doi = {https://doi.org/10.1016/j.epsr.2020.106547},
url = {https://www.sciencedirect.com/science/article/pii/S0378779620303515},
author = {Balthazar Donon and Rémy Clément and Benjamin Donnot and Antoine Marot and Isabelle Guyon and Marc Schoenauer}
}

@article{nellikkath2022,
title = {Physics-Informed Neural Networks for {AC} Optimal Power Flow},
journal = {Electric Power Systems Research},
volume = {212},
pages = {108412},
year = {2022},
issn = {0378-7796},
doi = {https://doi.org/10.1016/j.epsr.2022.108412},
url = {https://www.sciencedirect.com/science/article/pii/S0378779622005636},
author = {Rahul Nellikkath and Spyros Chatzivasileiadis}
}

@INPROCEEDINGS{opflearn,
  author={Joswig-Jones, Trager and Baker, Kyri and Zamzam, Ahmed S.},
  booktitle={2022 IEEE Power \& Energy Society Innovative Smart Grid Technologies Conference (ISGT)}, 
  title={{OPF-Learn}: {A}n Open-Source Framework for Creating Representative {AC} Optimal Power Flow Datasets}, 
  year={2022},
  volume={},
  number={},
  pages={1-5},
  doi={10.1109/ISGT50606.2022.9817509}
}

@misc{opfdata,
      title={{OPFData}: {L}arge-scale datasets for AC optimal power flow with topological perturbations},
      author={Sean Lovett and Miha Zgubic and Sofia Liguori and Sephora Madjiheurem and Hamish Tomlinson and Sophie Elster and Chris Apps and Sims Witherspoon and Luis Piloto},
      year={2024},
      eprint={2406.07234},
      archivePrefix={arXiv},
      primaryClass={cs.LG},
      url={https://arxiv.org/abs/2406.07234}, 
}

@inproceedings{pfdelta,
title={{PF}\ensuremath{\Delta}: A Benchmark Dataset for Power Flow under Load, Generation, and Topology Variations},
author={Ana K. Rivera and Anvita Bhagavathula and Alvaro Carbonero and Priya L. Donti},
booktitle={The Thirty-ninth Annual Conference on Neural Information Processing Systems Datasets and Benchmarks Track},
year={2025},
url={https://openreview.net/forum?id=Gi1HtsTAkv}
}

@inproceedings{szegedy2014intriguing,
  title={Intriguing properties of neural networks},
  author={Szegedy, Christian and Zaremba, Wojciech and Sutskever, Ilya and Bruna, Joan and Erhan, Dumitru and Goodfellow, Ian and Fergus, Rob},
  booktitle={International Conference on Learning Representations},
  year={2014},
  url={http://arxiv.org/abs/1312.6199}
}

@misc{pglib,
      title={The Power Grid Library for Benchmarking {AC} Optimal Power Flow Algorithms},
      author={Sogol Babaeinejadsarookolaee and Adam Birchfield and Richard D. Christie and Carleton Coffrin and Christopher DeMarco and Ruisheng Diao and Michael Ferris and Stephane Fliscounakis and Scott Greene and Renke Huang and Cedric Josz and Roman Korab and Bernard Lesieutre and Jean Maeght and Terrence W. K. Mak and Daniel K. Molzahn and Thomas J. Overbye and Patrick Panciatici and Byungkwon Park and Jonathan Snodgrass and Ahmad Tbaileh and Pascal Van Hentenryck and Ray Zimmerman},
      year={2021},
      eprint={1908.02788},
      archivePrefix={arXiv},
      primaryClass={math.OC},
      url={https://arxiv.org/abs/1908.02788}, 
}

@inproceedings{coffrin2018powermodels,
  author = {Carleton Coffrin and Russell Bent and Kaarthik Sundar and Yeesian Ng and Miles Lubin},
  title = {{PowerModels.jl}: {An} Open-Source Framework for Exploring Power Flow Formulations},
  booktitle = {2018 Power Systems Computation Conference (PSCC)},
  year = {2018},
  month = {June},
  pages = {1-8},
  doi = {10.23919/PSCC.2018.8442948}
}

@article{jump1,
author = {Lubin, Miles and Dowson, Oscar and Garcia, Joaquim Dias and Huchette, Joey and Legat, Beno{\^\i}t and Vielma, Juan Pablo},
date = {2023/09/01},
doi = {10.1007/s12532-023-00239-3},
id = {Lubin2023},
isbn = {1867-2957},
journal = {Mathematical Programming Computation},
number = {3},
pages = {581--589},
title = {{JuMP} 1.0: {R}ecent improvements to a modeling language for mathematical optimization},
url = {https://doi.org/10.1007/s12532-023-00239-3},
volume = {15},
year = {2023}
}

@article{parker2023dulmage,
title = {Applications of the {Dulmage-Mendelsohn} decomposition for debugging nonlinear optimization problems},
journal = {Computers \& Chemical Engineering},
volume = {178},
pages = {108383},
year = {2023},
issn = {0098-1354},
doi = {https://doi.org/10.1016/j.compchemeng.2023.108383},
url = {https://www.sciencedirect.com/science/article/pii/S0098135423002533},
author = {Robert B. Parker and Bethany L. Nicholson and John D. Siirola and Lorenz T. Biegler},
}

@misc{mathoptai,
      title={{MathOptAI.jl}: {Embed} trained machine learning predictors into {JuMP} models}, 
      author={Oscar Dowson and Robert B Parker and Russel Bent},
      year={2025},
      eprint={2507.03159},
      archivePrefix={arXiv},
      primaryClass={cs.LG},
      url={https://arxiv.org/abs/2507.03159}, 
}

@article{paszke2019pytorch,
  title={Pytorch: An imperative style, high-performance deep learning library},
  author={Paszke, Adam and Gross, Sam and Massa, Francisco and Lerer, Adam and Bradbury, James and Chanan, Gregory and Killeen, Trevor and Lin, Zeming and Gimelshein, Natalia and Antiga, Luca and others},
  journal={Advances in neural information processing systems},
  volume={32},
  year={2019}
}

@inproceedings{parker2025gpu,
title={Nonlinear Optimization with {GPU}-Accelerated Neural Network Constraints},
author={Robert B. Parker and Oscar Dowson and Nicole LoGiudice and Manuel J Garcia and Russell Bent},
booktitle={NeurIPS Workshop on GPU-Accelerated and Scalable Optimization},
year={2025},
url={https://openreview.net/forum?id=7Nf40tUjxL}
}

@article{ipopt,
  title={On the implementation of an interior-point filter line-search algorithm for large-scale nonlinear programming},
  author={W{\"a}chter, Andreas and Biegler, Lorenz T},
  journal={Mathematical programming},
  volume={106},
  number={1},
  pages={25--57},
  year={2006},
  publisher={Springer},
}

@article{ma57,
  author = {Duff, Iain S.},
  title = {{MA57}---a code for the solution of sparse symmetric definite and indefinite systems},
  year = {2004},
  issue_date = {June 2004},
  publisher = {Association for Computing Machinery},
  address = {New York, NY, USA},
  volume = {30},
  number = {2},
  issn = {0098-3500},
  doi = {10.1145/992200.992202},
}

@article{fioretto2020, title={Predicting {AC} Optimal Power Flows: {Combining} Deep Learning and Lagrangian Dual Methods}, volume={34}, url={https://ojs.aaai.org/index.php/AAAI/article/view/5403}, DOI={10.1609/aaai.v34i01.5403}, abstractNote={&lt;p&gt;The Optimal Power Flow (OPF) problem is a fundamental building block for the optimization of electrical power systems. It is nonlinear and nonconvex and computes the generator setpoints for power and voltage, given a set of load demands. It is often solved repeatedly under various conditions, either in real-time or in large-scale studies. This need is further exacerbated by the increasing stochasticity of power systems due to renewable energy sources in front and behind the meter. To address these challenges, this paper presents a deep learning approach to the OPF. The learning model exploits the information available in the similar states of the system (which is commonly available in practical applications), as well as a dual Lagrangian method to satisfy the physical and engineering constraints present in the OPF. The proposed model is evaluated on a large collection of realistic medium-sized power systems. The experimental results show that its predictions are highly accurate with average errors as low as 0.2%. Additionally, the proposed approach is shown to improve the accuracy of the widely adopted linear DC approximation by at least two orders of magnitude.&lt;/p&gt;}, number={01}, journal={Proceedings of the AAAI Conference on Artificial Intelligence}, author={Fioretto, Ferdinando and Mak, Terrence W.K. and Van Hentenryck, Pascal}, year={2020}, month={Apr.}, pages={630-637} }

@misc{powerplots,
      title={{PowerPlots.jl}: {An} Open Source Power Grid Visualization and Data Analysis Framework for Academic Research}, 
      author={Noah Rhodes},
      year={2025},
      eprint={2510.05063},
      archivePrefix={arXiv},
      primaryClass={eess.SY},
      url={https://arxiv.org/abs/2510.05063}, 
}

@misc{wiyatno2019adversarial,
      title={Adversarial Examples in Modern Machine Learning: {A} Review}, 
      author={Rey Reza Wiyatno and Anqi Xu and Ousmane Dia and Archy de Berker},
      year={2019},
      eprint={1911.05268},
      archivePrefix={arXiv},
      primaryClass={cs.LG},
      url={https://arxiv.org/abs/1911.05268}, 
}

@INPROCEEDINGS{chen2018vulnerable,
  author={Chen, Yize and Tan, Yushi and Deka, Deepjyoti},
  booktitle={2018 IEEE International Conference on Communications, Control, and Computing Technologies for Smart Grids (SmartGridComm)}, 
  title={Is Machine Learning in Power Systems Vulnerable?}, 
  year={2018},
  volume={},
  number={},
  pages={1-6},
  doi={10.1109/SmartGridComm.2018.8587547}}

@INPROCEEDINGS{dinh2023reliability,
  author={Dinh, My H. and Fioretto, Ferdinando and Mohammadian, Mostafa and Baker, Kyri},
  booktitle={2023 IEEE PES Innovative Smart Grid Technologies Latin America (ISGT-LA)}, 
  title={An Analysis of the Reliability of {AC} Optimal Power Flow Deep Learning Proxies}, 
  year={2023},
  volume={},
  number={},
  pages={170-174},
  doi={10.1109/ISGT-LA56058.2023.10328223}}

@ARTICLE{chevalier2026adversarial,
  author={Chevalier, Samuel and Wheeler, William A.},
  journal={IEEE Transactions on Power Systems}, 
  title={Identifying the Smallest Adversarial Load Perturbation that Renders DC-OPF Infeasible}, 
  year={2026},
  volume={},
  number={},
  pages={1-12},
  doi={10.1109/TPWRS.2026.3651838}}

@misc{chevalier2025maximal,
      title={Maximal Load Shedding Verification for Neural Network Models of AC Line Switching}, 
      author={Samuel Chevalier and Duncan Starkenburg and Robert Parker and Noah Rhodes},
      year={2025},
      eprint={2510.23806},
      archivePrefix={arXiv},
      primaryClass={eess.SY},
      url={https://arxiv.org/abs/2510.23806}, 
}

@misc{giraud2025training,
      title={Neural Networks for {AC} Optimal Power Flow: {Improving} Worst-Case Guarantees during Training}, 
      author={Bastien Giraud and Rahul Nellikath and Johanna Vorwerk and Maad Alowaifeer and Spyros Chatzivasileiadis},
      year={2025},
      eprint={2510.23196},
      archivePrefix={arXiv},
      primaryClass={eess.SY},
      url={https://arxiv.org/abs/2510.23196}, 
}

@inproceedings{donti2021dc,
title={{DC}3: {A} learning method for optimization with hard constraints},
author={Priya L. Donti and David Rolnick and J Zico Kolter},
booktitle={International Conference on Learning Representations},
year={2021},
url={https://openreview.net/forum?id=V1ZHVxJ6dSS}
}
\balance

\endgroup
\end{document}